\documentclass{bmvc}

\usepackage{graphicx}   

\usepackage{pragyansymb, subeqn} 
\usepackage[hang]{subfigure}

\begin{document}

\title{EigenFairing: 3D Model Fairing \\using Image Coherence}

%
%
\author{Pragyana Mishra, Omead Amidi, and Takeo Kanade\\
Robotics Institute, Carnegie Mellon University\\
Pittsburgh, Pennsylvania 15213, USA }

\maketitle

\begin{abstract}
A surface is often modeled as a triangulated mesh of 3D points and
textures associated with faces of the mesh. The 3D points could be
either sampled from range data or derived from a set of images using
a stereo or Structure-from-Motion algorithm. 
When the points do not lie at critical points of maximum curvature 
or discontinuities of the real surface, faces of the mesh do not 
lie close to the modeled surface. This results in textural 
artifacts, and the model is 
not perfectly coherent with a set of actual images---the ones that are 
used to texture-map its mesh. 

This paper presents a technique for perfecting the 3D surface model 
by repositioning its vertices so that it is coherent with a set of 
observed images of the object. 
The textural artifacts and incoherence with images are due to the 
non-planarity of a surface patch being approximated by a planar face, as 
observed from multiple viewpoints. Image areas from the viewpoints 
are used to represent texture for the patch in eigenspace. The eigenspace
representation captures variations of texture, which we seek to
minimize. 

A coherence measure based on the difference between the face 
textures reconstructed from eigenspace and the actual images is used 
to reposition the vertices so that the model is improved or faired. 
We refer to this technique of model refinement as EigenFairing,
by which the model is faired, both geometrically and texturally, to 
better approximate the real surface.

\end{abstract}

\section{Introduction}
\begin{figure}
\begin{center}
        \includegraphics[width=6.9cm]{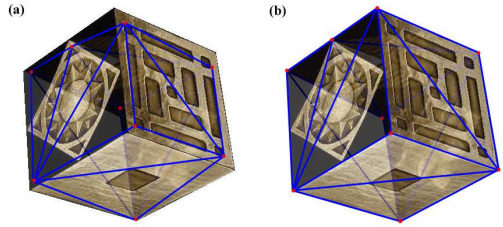}
\end{center}
 	\caption{(a) 3D points used for creating the model do not lie at geometrical extremities or points of 
	discontinuity of the cube's surface. Faces of the mesh with such 3D points as vertices are poor 
	approximation of the real surface. (b) For the new vertex positions, the mesh faces lie on the surface of the cube, thus creating a good model.} 
        \label{fig:EigFIntro01}
\end{figure}

A model of a real surface is often represented as a textured mesh of 3D points. 
The points form vertices of a triangulated mesh, faces of which are texture-mapped to model the surface.
This representation has proven to be effective, as a significant number of surfaces can be 
modeled by a mesh of planar faces.
Most sensed 3D data, however, do not always lie at critical points of maximum curvature or discontinuities, 
such as corners and edges of the real surface [Figure \ref{fig:EigFIntro01}].
Laser scanned data misses out on these strategic but spatially minute points. 
Features selected and tracked using local pixel intensities in images for stereo or SfM algorithms do not 
always correspond to points of discontinuity of the observed surface. 
These facts lead to a poor approximation of the surface by planar faces, and result in not-so-perfect
appearance of modeled surfaces. This paper brings out this salient aspect of 3D modeling. It presents a
new method of repositioning vertices of the mesh to those critical points for an improvement in the geometrical
and textural quality of the model. We show that this seemingly incremental change has a significant 
impact on the model's overall appearance. 

\begin{figure}
\begin{center}
        \includegraphics[width=12.5cm]{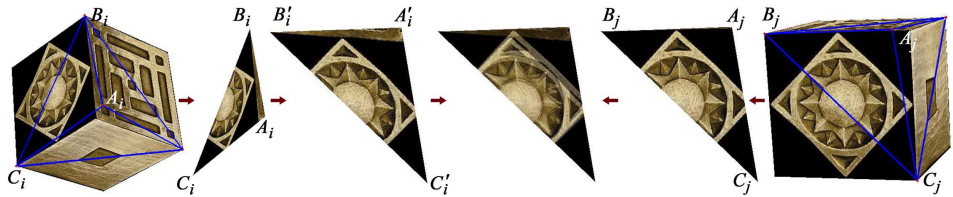}
\end{center}
        \caption{Image areas $A_iB_iC_i$ in the $i$-th image and $A_jB_jC_j$ in the $j$-th image 
correspond to the same 3D mesh face. Affine warping of $A_iB_iC_i$ into $A_jB_jC_j$ yields image 
patch $A'_iB'_jC'_j$. The texture within the two areas, $A'_iB'_iC'_i$ and $A_jB_jC_j$, 
are not the same because vertex $A$ is not at an edge or a corner, and the mesh face does not lie on the
actual 3D surface of the object. Note the ghosting of edges in their superimposition (average) in the center.}
        \label{fig:EigFInitPatch}
\end{figure}

The 3D model is refined or \textit{faired} by relocating vertices such that the planar faces 
lie closer to the surface patches they approximate. As the real surface is not 
known to us, the same set of images that is used for texture-mapping faces of the mesh 
guides the fairing process. 
A mesh face, that approximates a surface patch, is projected into images to yield
triangular image areas. A set of these image areas corresponding to the same surface patch has 
information of how good its planar approximation is. 
Had the surface patch been planar, the image areas corresponding to its planar approximation would 
be the same when warped into a fixed triangular area [Figure \ref{fig:EigFInitPatch}].
As the real world is not built of simplicial elements, the variations of texture in the image areas can be 
attributed to the non-planarity of the surface patch. 
We would like to get the best possible representation of texture from the image areas by encoding the textural
variations. 

Given a set of image areas, a compact representation of these variations can be derived in terms
of a small number of orthogonal basis images. This representation, also called an eigenspace decomposition,
encodes the minute variations of texture as observed from different viewpoints \cite{nishino91}.
We reposition vertices of the mesh such that these textural variations are minimized. 
In the process, faces of the mesh better approximate the real surface in the geometrical sense. 
Also, the texture-mapped 3D mesh--the 3D model--appears to be as close as possible to the 
actual images when observed from the same camera viewpoints.
This \textit{coherence} between the actual image area and the textural representation of the corresponding 
surface patch can be quantified as the distance-from-eigenspace of the image area. 
The better a triangular face approximates the surface patch geometrically, the smaller
is the variation of appearance in images, and the smaller is the distance-from-eigenspace
of the image-areas corresponding to the patch. We refer to this idea of 3D model refinement as 
\textit{EigenFairing}.   
\section{Related Work}
The problem of model refinement based on image data has been studied extensively.
A single image can be used to reposition vertices of a texture-mapped mesh by minimizing error
metrics based on the vertex positions, surface normals, and other surface properties such as
color and texture \cite{garland98}. The algorithm never accesses the pixels of the 
single-image texture and merely updates texture coordinates in the image.     
Image-driven mesh simplification \cite{lindstrom00} compares actual images
against images of the simplified model to decide which portions of a model to refine through the edge 
collapse operator. In image-consistent surface triangulation \cite{morris00}, the initial mesh
is refined through edge-swaps to best account for observed images. 
The edge-swapping scheme, that uses the weighted-average of affine-warped image areas, works well only
if a surface patch has enough texture and is close to the planar face that approximates it.     
Other algorithms \cite{FuaLecrec96cviu} \cite{NobuMatsu2003} recover surface 
shape and reflectance properties from
multiple images by deforming a 3D representation. However, these methods optimize complicated 
objective functions that combine several image and geometric-based constraints. We shall show that a
simple objective function, based on the eigenspace representation of texture, can be minimized
to refine the geometry of 3D models. Our approach is especially consequential for modeling surfaces 
whose textural detail is denser than the geometrical level of detail.     

Eigenspace texture methods \cite{nishino91} \cite{CootesEdTaylor2001} encode appearance variations of a
surface patch under various viewing conditions. Appearance of patches are
encoded using the eigenspace method, and new views are reconstructed from their
eigenspace representations. We shall use an eigenspace representation of dimensionality five 
to represent texture of 3D surface. It has been shown that for 
diffuse surfaces of arbitrary texture, the first five components of eigenspace explain most
of the image variation \cite{epstein95}.

EigenFairing is similar to the multiresolution surface reconstruction algorithm \cite{ZhangSeitz2000},
but it does not attempt to subdivide the mesh to account for perspective effects. It refines a
given mesh to best approximate an observed surface. The planarity of a surface patch as compared 
to a mesh face that models it, i.e. perspective distortion, depends on the position of the vertices. 
One of the critical components of model refinement, that seems to be missing in
previous work, is the relocation of vertex points so that they lie at the extremities or critical
points of object geometry.  
\section{Image Coherence}   
If a surface patch is not planar and has perspective distortion, how can texture of its 
planar approximation---the mesh face---be best represented, given the triangular pixel-areas in multiple 
images that correspond to it? The triangular image areas depend on viewpoints of the 
camera as well as degree of non-planarity of the 3D-surface patch. 
The affine warping of triangular image areas into each other does not align the texture 
within [Figure \ref{fig:EigFInitPatch}]. 
Estimating texture that corresponds to the mesh face
by weighted-average of pixels from different images \cite{morris00} leads to blurring of the
estimated texture. Since image-consistency is sensitive to textural variations, averaging of 
affine-warped image areas leads to complicated overlapping surfaces and chances of the refinement
algorithm being trapped in local minima. Moreover, textural variations of image areas corresponding
to the same 3D-surface patch can be exploited for refining the mesh. Image coherence considers
this \textit{intra-image} textural variation, as well as consistency of estimated texture when 
compared to actual images.

For a given set of image areas corresponding to a single 3D-surface patch, image coherence constructs
a small set of basis images that best captures the variations in texture. 
These basis images form a view-based representation of the texture of the patch. 
Each triangular image area is affine warped to a fixed triangular area called the \textit{cell image} 
\cite{nishino91}. 
Since each cell image is created by affine warping the triangular image areas corresponding to
a patch, the number of basis vectors to adequately represent its appearance depends on the planarity
of the patch. This is valid under the assumption that the patch has Lambertian reflective 
properties. The more planar a patch is, the better is its approximation by the mesh face, the
lower is its perspective distortion as seen from the image-viewpoints, and therefore, the lower 
is the number of basis vectors needed to represent the texture of the patch.
Considering the first $k$ principal components corresponding to the $k$ largest singular 
values, the basis image-set is $\mathbi{U}_\cf = [U_{\cf 1}, U_{\cf 2}, ..., U_{\cf k}]$ for face $\cf$.
Let $U_\cf \bc_i$ denote the reconstructed cell image corresponding to the $i^{th}$-image patch.
The set of scalar values, $ \bc_i = [c_{i1}, c_{i2}, ..., c_{ik}]$, is computed by taking the dot product
of the cell image and basis images $\mathbi{U}_\cf$. Thus, $U_\cf \bc_i$ is constructed by
a linear combination of the $k$ basis images; $U_\cf \bc_i \equiv \sum_{m=1}^k  c_{im} U_{\cf m}$. 
\begin{figure}
\begin{center}
        \includegraphics[width=12.5cm]{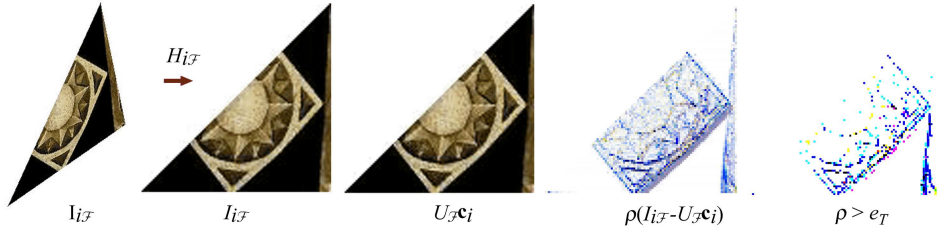}
\end{center}
        \caption{ The triangular image patch ${\rm I}_{i \cf}$ in the $i$-th image corresponding to face 
	$\cf$ is warped to a fixed cell image $I_{i \cf}$. 
	The reconstructed cell image using the first five components of eigen-space is $U_\cf\bc_i$. 
	The Geman-McClure norm between  $I_{i \cf}$ and $U_\cf\bc_i$ for $\sigma$=100, and the spatial distribution of outliers ($e_T = \sigma /\sqrt{3}$) are on the right.}  
        \label{fig:EigFCell}
\end{figure}

An image coherent representation is the best possible approximation, $U_\cf \bc_i$, over all image
areas, i.e., for $i= 1, 2, ..., n$. For a Lambertian surface, such a representation reflects
the accuracy of approximation of the textured patch by the mesh geometrical element. Image
coherence is based on the closeness of these two approximations: the textural
approximation as captured by the basis images, and the geometrical approximation of the physical
3D patch by the planar face. Minimizing the error in representation of surface texture minimizes
the error in the approximation of the surface patch by a linear mesh element.

\section{Model Fairing}
Given a set of basis texture-images $\mathbi{U}_\cf$ corresponding to
a face $\cf$, we reconstruct the image patch $U_\cf \bc_i$ in the $i$-th image. 
The objective function to be minimized for the set of $n$ cell images, 
$I_{i \cf} \:$; $i=1, ..., n$ , is 
\begin{equation}
E(\bc) = \sum_{i=1}^n \sum_{\biu_\cf} \rho ( I_{i \cf}(\biu) - [U_\cf \bc_i](\biu))
\end{equation}
The error norm $\rho$ defined over residual pixel-error in cell images.
$\biu_\cf$ is the coordinate of cell pixels.
Instead of exhaustively searching around each vertex in 3D space, we formulate
an iterative search method.   
\subsection{Vertex Displacement}
Let $\bfeta=\left[ \eta_X \; \eta_Y \; \eta_Z \right]^T$ represent a 3D displacement of a vertex. The faired
vertex point $\tilde{\bx}_\cv$ is updated as $\tilde{\bx}_\cv \leftarrow \bx_\cv + \bfeta$.
The goal is to simultaneously find the coefficients $\bc$ and displacement vector $\bfeta$ 
that minimize the objective function of the residual error;
\begin{equation}
E( \bc, \bfeta) = \sum_{i=1}^n \sum_{\biu_{\tilde{\cf}}} \rho (I_{i \tilde{\cf}}(\biu) - [U_{\tilde{\cf}} \bc_i](\biu))
\label{eqn:objfn}
\end{equation} 
for the faces $\tilde{\cf}$ constructed from the new vertices $\tilde{\bx}_\cv$. 
This optimization interleaves two sub-problems. The first sub-problem
is to minimize $E(\bc, \bfeta)$ with respect to $\bc$ while the vertex $\bx_\cv$ is
kept fixed. This is the same as the eigen-texture method \cite{nishino91} discussed in the last section.

The second sub-problem is to minimize $E(\bc, \bfeta)$ with respect to the \textit{fairing} 
parameters or displacement $\bfeta$, this time with the coefficients $\bc$ held fixed. 
The image patch $I_{\cf}$ corresponding to face $\cf$ gets warped to $I_{\tilde{\cf}}$
that corresponds to a new face $\tilde{\cf}$. For a given set of basis images 
$\mathbi{U}_\cf$, we have to determine a new set of image patches
and corresponding cell images, $I_{i \tilde{\cf}}(\biu)$ for $i=1, 2, ..., n$ , such that the following is minimized:
\begin{equation}
\sum_{i=1}^n \sum_{\biu_{\tilde{\cf}}} \rho (I_{i \tilde{\cf}}(\biu) - [U_\cf \bc_i](\biu))
\label{eqn:objfn2}
\end{equation} 
Due to a vertex displacement $\bfeta$, cell pixels at $\biu$ get displaced to new cell 
image coordinates as $\biu + \upsilon_{i \cf}(\biu, \bfeta)$. 
For a given pixel-coordinate displacement function $\upsilon_{i \cf}(\biu, \bfeta)$, the new cell image is  
$I_{i \tilde{\cf}}(\biu) = I_{i \cf}(\biu + \upsilon_{i \cf}(\biu, \bfeta))$. 
Ideally, we should have 
\begin{equation}
I_{i \cf}(\biu + \upsilon_{i \cf}(\biu, \bfeta)) = [U_{\cf} \bc_i](\biu)
\label{eqn:param1}
\end{equation}
Equation (\ref{eqn:param1}) states that there are pixel displacements 
$\upsilon_{i \cf}(\biu, \bfeta)$ that
when applied to the image patch $I_{i \cf}$ make $I_{i \cf}$ look like some image reconstructed
from the eigenspace. 
A first order Taylor series expansion of the left hand side of Equation (\ref{eqn:param1}) yields
\begin{equation}
I_{i \cf}(\biu) + \nabla I_{i \cf} \cdot \upsilon_{i \cf}(\biu, \bfeta) = [U_{\cf} \bc_i](\biu)
\label{eqn:param2}
\end{equation}
Summing the residual pixel-error $e_c$ over cell pixels $\biu_\cf$ corresponding to face $\cf$ in all images, 
the error function can be written in terms of $e_c$ as 
\begin{equation}
E( \bc, \bfeta) = \sum_{i=1}^n \sum_{\biu_\cf} \rho \left( \: e_c \: \right)
\hspace{0.5cm} {\rm where} \hspace{0.5cm}
e_c \equiv \nabla I_{i \cf} \cdot \upsilon_{i \cf}(\biu, \bfeta) + \left( I_{i \cf}(\biu) - [U_{\cf} \bc_i](\biu) \right) 
\label{eqn:param3}
\end{equation} 
\subsection{Optimization}
The minimization of $E(\bc, \bfeta)$ with respect to $\bfeta$ can be obtained using the 
Gauss-Newton algorithm.
In the Gauss-Newton method, a search direction is computed using the gradient, and a first-order 
approximation to the Hessian for the given objective function $E(\bc, \bfeta)$. 
The $k$-th element of gradient vector $\bg$ is
\begin{equation}
\bg_k = 
\sum_{i=1}^n \sum_{\biu_\cf} 
\dot{\rho} \left( \: e_c \: \right) \:
\frac{\partial e_c}{\partial \eta_k}
\hspace{0.25cm} {\rm ,} \hspace{0.5cm} 
\frac{\partial e_c}{\partial \eta_k} = 
\left( \nabla I_{i \cf} \cdot \frac{\partial \upsilon_{i \cf}(\biu, \bfeta)}{\partial \eta_k} \right)
\hspace{0.5cm} {\rm for} \hspace{0.5cm} k \in \{ X, Y, Z\}.
\end{equation}
where $\dot{\rho}$, also called the influence function, is the derivative of the error norm $\rho$ 
with respect to the residual pixel error.
We have chosen the Geman-McClure norm as our error norm $\rho$; it is defined over the residual 
pixel-error $e_c$ in cell images. 
Given a scale factor $\sigma$ that controls the convexity of the norm and its influence to outliers, we have:
$ \rho(e_c) = e_c^2 / (\sigma+e_c^2) \; \:$,  
$\dot{\rho}(e_c) = 2\sigma e_c / (\sigma+e_c^2)^2 \; \:$, and 
$\ddot{\rho}(e_c)= 2\sigma(\sigma-3e_c^2) / (\sigma+e_c^2)^3$.
The $\{k,l\}$-th element of the Hessian $\mathbf{H}$ is
\begin{equation}
\mathbf{H}_{kl} = \sum_{i=1}^n \sum_{\biu_\cf}
\ddot{\rho}(e_c)
\frac{\partial e_c}{\partial \eta_k}
\frac{\partial e_c}{\partial \eta_l} 
\hspace{0.5cm} {\rm for} \hspace{0.5cm} k,l \in \{X, Y, Z \}
\label{eqn:hess01}
\end{equation}
The objective function $E$ is convex when the Hessian $\mathbf{H}$ of $E$ is positive definite.
A positive definite Hessian indicates that the function has a unique optimum, in the local neighborhood, 
whereas a Hessian that has one or more eigenvalues zero will allow an entire manifold of solutions
to minimize the objective function.  
$E$ is locally convex when $\ddot{\rho}(e_c) > 0 \; \forall \; e_c$.
For small $\sigma$ values, 
$\ddot{\rho}(e_c)$ can be negative, and therefore, one may not get a descent direction. 
As $\dot{\rho}(0)=0 \:$, $\ddot{\rho}(e_c)$ is substituted by its secant approximation, $\dot{\rho}(e_c)/e_c$, 
for small values of $e_c$, and is positive everywhere. Substituting in equation \ref{eqn:hess01}, 
\begin{eqnarray}
\sum_{l \in \{ X, Y, Z \}} 
\left( \sum_{i=1}^n \sum_{\biu_\cf} \frac{\dot{\rho}(e_c)}{e_c} 
\frac{\partial e_c}{\partial \eta_k} \frac{\partial e_c}{\partial \eta_l} \right)
\delta \eta_l =  
-\sum_{i=1}^n \sum_{\biu_\cf}  
\dot{\rho}(e_c) 
\frac{\partial e_c}{\partial \eta_k} 
\hspace{0.25cm}{\rm for}\hspace{0.25cm} k \in \{ X, Y, Z\}.
\end{eqnarray}
The fairing displacements, $\bfeta=\left[ \eta_X \; \eta_Y \; \eta_Z \right]^T$, are iteratively updated for step $\: m \:$ as $\bfeta^{(m+1)} = \bfeta^{(m)} + \delta \bfeta$
and $\bfeta^{(0)} = \left[\:\: 0 \;\: 0 \;\: 0 \:\:\right]^T$. 
After each vertex \textit{fairing} step, the image patches are updated by considering
the new position of the vertex. The new image patches are warped into their corresponding cell images. 
These cell images are used for the next optimization step. As the warping registers the image patches
and the eigenspace, the approximation $[U_{\cf} \bc_i]$ continues to improve and the texture representation
gets better. 
\begin{figure}[t!]
\begin{center}
        \includegraphics[width=12.5cm]{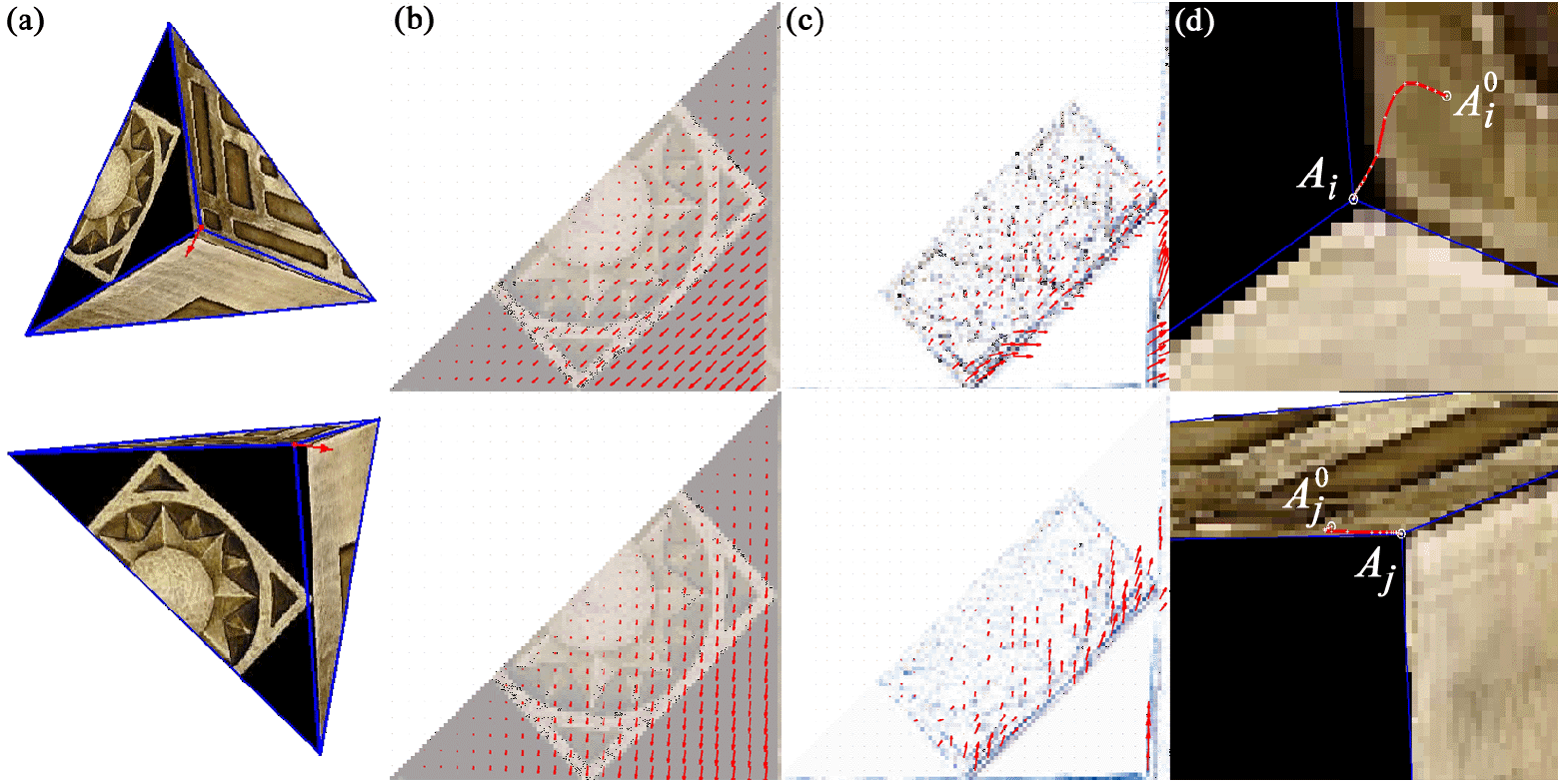}
\end{center}
        \caption{(a) A vertex displacement induces pixel displacements $\delta \bu_{i \cv}$ and $\delta \bu_{i \cv}$, 
shown as red arrows, in image $i$ and $j. \;$ (b) Pixel displacements are used to calculate the cell pixel-displacements 
$\upsilon_{i \cf}$ and $\upsilon_{j \cf}$ for the $i$-th and $j$-th cell images. 
(c) Optical flow calculated using constant brightness constraint between the cell image and its eigenspace reconstruction is superimposed on their difference. (d) The path of successive vertex displacements. $A^0_i$ and $A^0_j$ correspond to the initial position of the vertex. $A_i$ and $A_j$ correspond to the 3D point that the vertex converges to. (All red arrows are displacements magnified 10$\times$)}
        \label{fig:EigFFlowPath}
\end{figure} 
\subsection{Cell-Pixel Displacement}
The cell-pixel displacement function $\upsilon_{i \cf}(\biu, \bfeta)$ in the $i$-th image, 
corresponding to the face $\cf$, is related to the vertex-displacement $\bfeta$. 
The $3 \times 4$ projection matrix $\bP$ for the $i$-th image is known, 
and let its $p$-th row be represented as $\;\left[ \; \br_p \; | \; t_p \; \right]$ for $\;p = 1,2$ and $3$.
The pixel displacement in the $i$-th image corresponding to a small 3D vertex displacement, $\bfeta$, is
\begin{equation}
\delta \bu_{i \cv} =
\left[ \begin{array}{cc} \frac{(\br_3 \bx + t_3)\br_1 \bfeta - (\br_1 \bx + t_1)\br_3 \bfeta}{(\br_3 \bx + t_3)^2} & 
\frac{(\br_3 \bx + t_3)\br_2 \bfeta - (\br_2 \bx + t_2)\br_3 \bfeta}{(\br_3 \bx + t_3)^2}  \end{array} \right]^T
\end{equation}     	
This pixel displacement causes a change in the pixels of the cells. For pixels corresponding to face $\cf$ in the $i$-th image,
the affine transformation $H_{i \cf}$ (a $2 \times 3$ matrix) between the image patch and its cell is known, and 
$\biu = H_{i \cf} \; [ \bu^T \; \; 1]^T$.
The cell-pixel displacement function can now be related to the vertex pixel displacement in the image as
\begin{equation}
\upsilon_{i \cf}(\biu_\cv, \bfeta) = H_{i \cf} \; [\delta \bu_{i \cv}^T \; \; 0]^T \; + \;
\delta H_{i \cf} \:\: [\bu_{i \cv}^T \; \; 1]^T   
\end{equation}
For small pixel displacements, $\delta H_{i \cf}$ is usually negligible. 
Once the cell-pixel displacement at the moving vertex is determined, cell-pixel displacements over the
entire cell are calculated by interpolation, because the displacements at the other two \textit{fixed}
vertices and along their connecting edge are zero. The goal is to achieve cell-pixel displacements that
vary linearly across the cell image from the moving vertex to the other two vertices, or are as close as possible 
to a linear variation.  
We are trying to determine the location of the \textit{faired} vertex that best approximates the surface
patch with a planar face of the mesh. The same vertex location yields observed image-pixel displacements that are 
closest to linear variation of image-pixel displacements of a planar face. 
For the new position of the vertex, the cell-pixel displacements
lead to a better approximation of the surface patch; hence, a \textit{more} linear variation of flow or pixel
displacements appears across the cell image.    

\section{An Illustration}

At first, we illustrate an EigenFairing process for a \textit{unit} cube with textures on 
three faces visible in $12$ images. 
The vertex $A$, common to the three faces as shown in Figure \ref{fig:EigFInitPatch},
is faired such that it corresponds to the actual corner of the cube. 
The cell size is $128 \times 128$ pixels. 
The path of the vertex, as it moves towards the actual corner,
is displayed in Figure \ref{fig:EigFFlowPath}(d) for two views. 
A vertex displacement during the fairing process is shown in Figure \ref{fig:EigFFlowPath}(a). 
The interpolated pixel-displacements in the cell images, for
the same vertex displacement, are shown in Figure \ref{fig:EigFFlowPath}(b). 
Figure \ref{fig:EigFFlowPath}(c) shows the optical
flow fields computed between the cell image and its eigenspace reconstruction. They point in 
the opposite direction to the pixel-displacements. The estimated vertex displacement is the change
in vertex position for which induced pixel-displacements on faces best counter-balance flow-fields
between the cell-images and their eigenspace reconstructions.    
\section{Results with Real Data}
\begin{figure}[t!]
      \includegraphics[width=12.5cm]{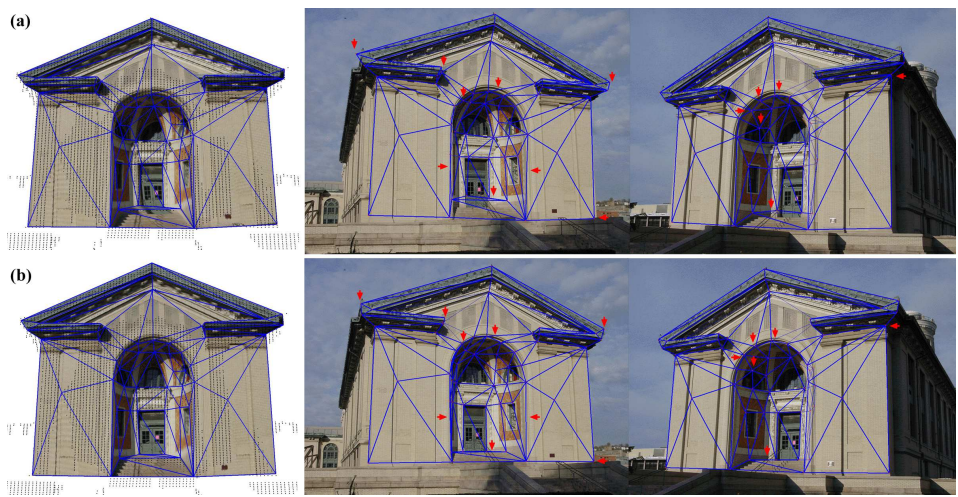}  
      \caption{On the first column, 3D range data is superimposed on the texture-mapped models before and after EigenFairing. Range scans often miss out corners due to occlusions or their spatial minuteness. 
Selected vertices are not always best positioned for surface approximation, as seen on the inner dome. The edges are close to, but not perfectly aligned with the physical edges of the building. Row (a) Initial mesh. Row (b) The EigenFaired mesh. The red arrows point at mesh elements that have considerably improved the model.}
        \label{fig:HamMesh}
\end{figure}

\begin{figure}[t!] 
	\includegraphics[width=12.5cm]{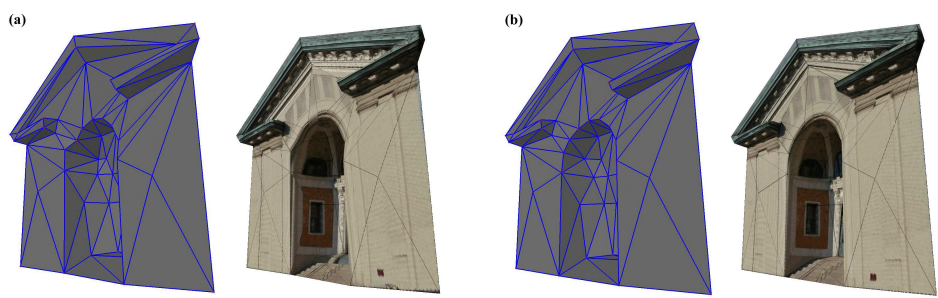}
	\caption{(a) The mesh and the texture-mapped model for the initial position of vertices. (b) The EigenFaired mesh and model. Notice that the initial mesh has irregularities in structure along the arch and on the inner dome. The corresponding textural artifacts are clearly visible in the model. The EigenFaired mesh better approximates the geometrical structure of the building. Its texture-mapped model is devoid of artifacts at the same time.}
        \label{fig:HamModel}
\end{figure}

\begin{figure} 
	\includegraphics[width=12.5cm]{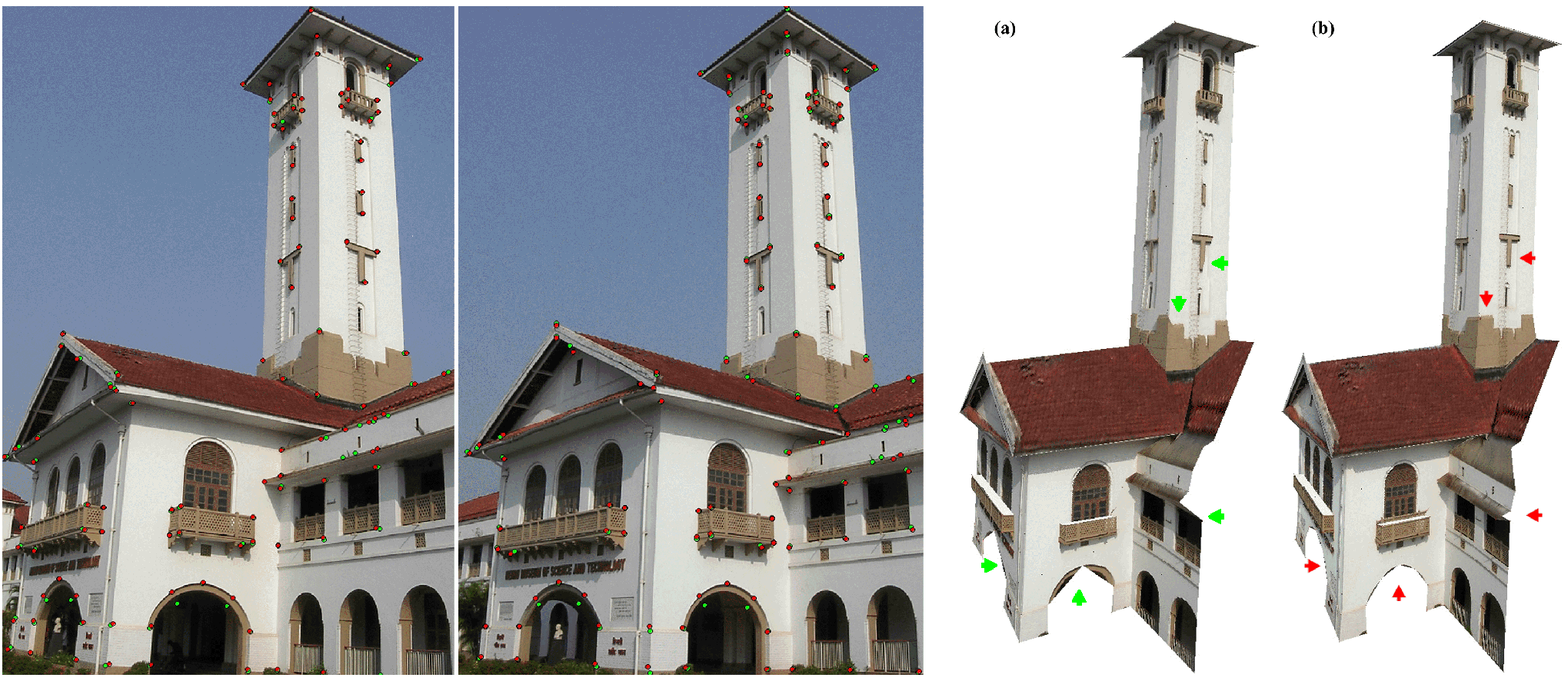}
	\caption{On the left are two images showing the initial vertex positions in green and the faired vertices in red. The faired vertices accurately correspond to corners and edges of the architecture. Notice the points along the archways, roof and balcony edges. The initial model and the EigenFaired model are shown in (a) and (b) respectively. The green and red arrows point out the presence and absence of modeling artifacts respectively.}
        \label{fig:IITModel}
\end{figure}

We applied EigenFairing to real outdoor scenes. 
An initial set of 3D points for generating the mesh was sampled from range 
data at points of geometrical extremities and discontinuities, as shown in Figure
\ref{fig:HamMesh}(a).
The data was registered with a sequence of 22 images, two of which are shown in Figure
\ref{fig:HamMesh}, by selecting the closest points of correspondence.  
The points do not accurately correspond to physical corners in images, and
edges of the mesh are not aligned with edges of the facade, as viewed in the images. Notice the
textural artifacts if this mesh is used for modeling the facade, in 
Figures \ref{fig:HamMesh}(a, left column) and \ref{fig:HamModel}(a). 
After EigenFaring the initial model, the 3D vertices correspond to actual corners, and 
edges of the mesh are aligned with physical edges in images [Figure \ref{fig:HamMesh}(b)]. 
Also, the facade is better modeled in a geometrical sense and, at the same time, 
devoid of textural artifacts [Figure \ref{fig:HamModel}(b)]. Figure \ref{fig:IITModel}
shows another real data set where a SfM algorithm was used to derive the initial position of
vertices from 10 images. For both data sets, the initial meshes are generated using 
Image-consistent Triangulation \cite{morris00}.  

The robustness of the fairing process was increased by adding a   
coarse-to-fine refinement strategy. A five-level pyramid was used where 
$\sigma_{smooth}$ varied linearly from 6.0 to 1.2. The Geman-McClure 
parameter was chosen as $\sigma = max(|e_c|)/\sqrt{3}$.
The error norm minimizes the effect of outliers that usually appear
at geometrical edges of image areas, specularitites, or due to large textural discontinuities
within a patch [Figure \ref{fig:EigFCell}].  
The neighborhood of 3D points for which the initial vertex converges to the faired position depends 
on how textured the neighboring faces are and how well
their textures are represented in eigenspace. The new 3D point may not always
lie on the actual surface, although its neighboring faces collectively represent 
the actual surface region with greater accuracy, both geometrically and 
texturally.

\section{Conclusion}
The EigenFairing algorithm refines a texture-mapped 
mesh to better represent a surface by relocating vertices of the mesh. The 
relocation process is carried out such that the texture mapped onto the
faces is best represented, and the resulting 3D model is coherent with 
observed images. Coherence of texture-mapped faces with images has been 
formulated as minimizing the distance between observed textures
corresponding to the faces of the mesh and their eigenspace reconstructions.
We have shown that minimizing this distance leads to a better geometrical
approximation of the unknown surface by the 3D model.     
EigenFairing couples geometrical properties of a 3D model with 
its textural properties or albedo. Coherence of texture leads to a 
faired mesh that is physically closer to the true surface. This approach is significant
for modeling a surface whose textural variation is much higher than the 
geometrical resolution required to represent the surface. 


\bibliographystyle{plain}
\bibliography{paperBib}

\end{document}